# Hardening DNNs against Transfer Attacks during Network Compression using Greedy Adversarial Pruning

**Jonah O'Brien Weiss**
Dept. of Electrical and Computer Engineering
University of Massachusetts Amherst
Amherst, MA
jobrienweiss@umass.edu

**Tiago Alves**
Dept. of Informatics and Computer Science
State University of Rio de Janeiro - UERJ
Rio de Janeiro, Brazil
tiago@ime.uerj.br

**Sandip Kundu**
Dept. of Electrical and Computer Engineering
University of Massachusetts Amherst
Amherst, MA
kundu@umass.edu

*Abstract—* The prevalence and success of Deep Neural Network (DNN) applications in recent years have motivated research on DNN compression, such as pruning and quantization. These techniques accelerate model inference, reduce power consumption, and reduce the size and complexity of the hardware necessary to run DNNs, all with little to no loss in accuracy. However, since DNNs are vulnerable to adversarial inputs, it is important to consider the relationship between compression and adversarial robustness. In this work, we investigate the adversarial robustness of models produced by several irregular pruning schemes and by 8-bit quantization. Additionally, while conventional pruning removes the least important parameters in a DNN, we investigate the effect of an unconventional pruning method: removing the most important model parameters based on the gradient on adversarial inputs. We call this method Greedy Adversarial Pruning (GAP) and we find that this pruning method results in models that are resistant to transfer attacks from their uncompressed counterparts. Code is available at [1].

## I. INTRODUCTION

Recent work has extensively proven the vulnerability of Deep Neural Networks (DNNs) to adversarial inputs, imperceptibly small perturbations added to a benign input that cause misclassification [2]. Adversarial inputs pose a legitimate concern for several reasons: they are not difficult for an attacker to find [2, 3], they often cause misclassification on multiple DNNs [2, 4], previous attempts to build countermeasures have failed [3, 5], and DNNs may be intrinsically vulnerable to them [2, 6]. Thus, the robustness of a DNN, or the amount noise which may be added to inputs without causing misclassification (commonly represented as an $L_2$ or $L_\infty$ norm) [3], has become an important point of consideration [2, 4, 5, 7-20]. Furthermore, with the rise of edge computing [21], DNNs are migrating to mobile and embedded devices, necessitating security measures in these environments [9].

Concurrent with the move to edge computing, but orthogonal to security, there have been several compression methods devised to adapt DNNs to resource constrained environments. Quantization [21] reduces the numerical precision of the network's parameters to lessen the memory footprint and reduce energy consumption since lower bitwidth operations require less power. Some quantization schemes allow for computation on the quantized data, which speeds up inference [21]. Pruning [22], on the other hand, is the process of removing the least important parameters of a network so that during inference, the operations which would have included those parameters can be skipped. The typical pruning pipeline involves training a model, pruning it, and then training it for a few more epochs in a process known as fine-tuning [23]. Pruning methods are usually classified based on two categories: their regularity (or irregularity), which provides constraints on the pattern of parameters to remove, and the function ranking the parameters' importance [23].

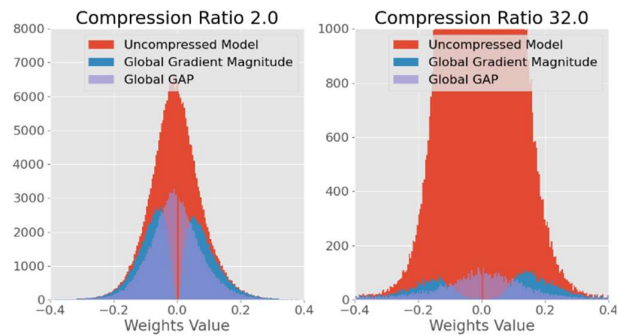

Fig. 1. Distribution of weight values remaining in a ResNet20 model after Greedy Adversarial Pruning (GAP) and Global Gradient Pruning for prune compression ratios of 2 and 32. These distributions are captured before fine-tuning to show the effect of the pruning operation, so the final pruned models will have slightly different distributions.

There is a significant amount of work related to the robustness of DNNs against adversarial examples [2, 3, 5, 6]. As a starting point, uncompressed DNNs with higher capacity are more robust than those with smaller capacity [3, 7]. Overparameterized networks may learn redundant or unneeded features [24], some of which may be useful for both clean and adversarial classification, while others may be useful just for clean classification [6]. Pruned models may achieve more robustness up to a certain sparsity, after which normal and adversarial accuracy decreases substantially [10, 11]. Adversarial inputs transfer from unpruned to pruned models [12], and from full precision to quantized models at high attack strengths due to an error amplification effect, but not low attack strengths, where the error may be attenuated [8]. However, there is less transferability between quantized models of different bitwidths [13], or models with different sparsity patterns [14].

There have also been many works seeking to combine compression methods with adversarial robustness in mind. These include simultaneous adversarial training and pruning [7, 15-17], simultaneous adversarial training and quantization [8, 19], or all three at once [9]. There have also been efforts to apply stochastic pruning [14] or quantization [13, 18] at inference time, although these methods do not provide much DNN performance optimization. Finally, there is one method proposed to prune one feature layer based on the difference between adversarial activations and clean activations, but without adversarial training [20].

**Our Contributions:** We investigate the adversarial robustness of several conventional irregular pruning methods, as well as their quantized counterparts, on state-of-the-art DNNs. We also introduce Greedy Adversarial Pruning (GAP), which prunes the parameters with the highest adversarial gradients in order to remove a DNN's ability to misclassify a known set of adversarial inputs. We demonstrate that although the DNNs produced from GAP are not robust to adversarial inputs generated on themselves, they are robust to transfer attacks with adversarial inputs generated from their uncompressed counterparts. These results suggest that the sparsity patterns induced by GAP can change the decision boundaries on a known set of adversarial examples.

## II. METHODOLOGY

As in [23], let $f(x;\cdot)$ represent an unpruned model architecture with uninitialized parameters. Then $f(x;\theta)$ represents a model, or an instantiation of a model architecture parameterized by $\theta$. For example, ResNet20 is a model architecture, while an instance of ResNet20 trained on CIFAR10 is a model. The pruning process produces a new model $f(x; M \odot \theta)$, where $M \in \{0, 1\}^{|\theta|}$ is a binary mask in which a 0 indicates a pruned parameter and $|\theta|$ denotes the cardinality of $\theta$. The pruned model is parameterized by the Hadamard product of the mask and the unpruned parameters.

For a prune compression $p$ defined as the ratio of the number of parameters in the original model to the number of parameters in the pruned model, $M$ is generated according to:

$$M_n = \begin{cases} 0, & s(\theta_n) < \gamma, \\ 1, & \text{else} \end{cases} \quad (1)$$

where the scoring function $s(\cdot)$ denotes the relative importance of each parameter $\theta_n$ and $\gamma$ is the $1 - 1/p$ percentile of importances. $M$ may be generated one layer at a time or globally over the entire network. For magnitude pruning, in which the parameters with the smallest absolute magnitude are removed, the scoring function is formulated as

$$s(\theta_n) = |\theta_n| \quad (2)$$

where $|\theta_n|$ denotes the absolute value of parameter $\theta_n$. For gradient pruning, in which the parameters with the smallest absolute gradient are removed, the scoring function is formulated as

$$s(\theta_n) = \sum_{(x,y) \in \mathcal{D}} |\nabla_{\theta_n} L(\theta, x, y)| \quad (3)$$

where $x$ is a datapoint in the training data $\mathcal{D}$ with label $y$ and $L(\theta, x, y)$ is the loss used to train the network.

Since first-order attacks use gradient ascent to generate adversarial inputs [3], we introduce Greedy Adversarial Pruning (GAP) which removes the parameters with the highest positive adversarial gradient. GAP is formalized by (1) by

$$s(\theta_n) = \sum_{(x,y) \in \mathcal{D}} -\nabla_{\theta_n} L(\theta, x_{adv}, y) \quad (4)$$

where $x_{adv}$ is an adversarial input generated from an unperturbed input $x$. The negative sign causes the highest adversarial gradients to be pruned, and by omitting the absolute value, the highest positive adversarial gradients are pruned, which is why this is a greedy method.

Adversarial inputs are generated using Projected Gradient Descent (PGD), which is a model of all first-order adversaries [3]. PGD is an iterative attack in which the adversarial image $x^{t+1}$ of the next round is calculated according to

$$x^{t+1} = \prod_{x+\epsilon}(x^t + \alpha \text{ sgn}(\nabla_x L(\theta, x, y))) \quad (5)$$

where the sgn($\cdot$) function returns the sign of the input, $\alpha$ is the step size, and the $\prod_{x+\epsilon}(\cdot)$ operator projects the adversarial image back into the $L_\infty$ bound of size $\epsilon$ around $x$.

## III. EXPERIMENTAL RESULTS AND DISCUSSION

We apply GAP layerwise and globally to four state-of-the-art architectures trained on CIFAR10: ResNet20, VGG, GoogLeNet, and MobileNetV2. We evaluate three metrics: test accuracy on unperturbed inputs from the test data, adversarial accuracy on adversarial inputs generated from the training data, and transfer attack accuracy, in which adversarial inputs generated from the training data on the unpruned model are applied to the pruned model. Additionally, to further investigate the adversarial robustness of irregularly pruned DNNs, we implement 5 conventional pruning methods from Shrinkbench [23] for comparison: global random pruning, global and layerwise magnitude pruning, and global and layerwise gradient pruning. As in [23], we do not prune the fully connected layer before the logits layer, we use single-shot, non-iterative pruning, and we evaluate the models at pruning compression ratios of 2, 4, 8, 16, and 32. Finally, to investigate the combined effect of pruning and quantization, we implement post-training quantization on all pruned and unpruned models and evaluate the three metrics described above: the test accuracy, the adversarial accuracy, and the transfer attack accuracy.

We use implementations of ResNet20 and VGG from Shrinkbench [23], GoogLeNet and MobileNetV2 from PyTorch, and implementation of PGD from CleverHans [25]. For training, we use the same hyperparameters as [7] for CIFAR10: training with SGD for 300 epochs with a learning rate of 0.1 that is divided by 10 at a quarter and halfway through training, and weight decay of 0.0001. Pruned models are fine-tuned for 40 epochs with a learning rate of 0.001. We use a batch size of 256 for training and fine-tuning. To generate adversarial inputs, we use PGD with random start, $\epsilon = 8/255$, $t = 10$, and $\alpha = 2/255$ as is common in the literature [7, 16]. To quantize models, we use PyTorch's post-training quantization to reduce parameter precision to 8 bits. Since the gradient of quantized models is non-differentiable, we use the gradient of their full-precision counterparts when generating adversarial inputs for quantized models.

Figure 2 shows the test accuracy, adversarial accuracy, and transfer attack accuracy for all models, pruning strategies, and compression ratios. In terms of test accuracy, we observe that layerwise application of GAP failed to learn the task except for ResNet20 and VGG at the compression ratio of 2. Random pruning suffers significant accuracy loss as compared to other methods, while in general, global pruning is more accurate than layerwise and pruning by parameter magnitude is more accurate than by gradient. Global GAP achieves slightly less accuracy in relation to conventional pruning methods, yet stills learns despite pruning more parameters with larger magnitude, and retains accuracy well as compression increases.

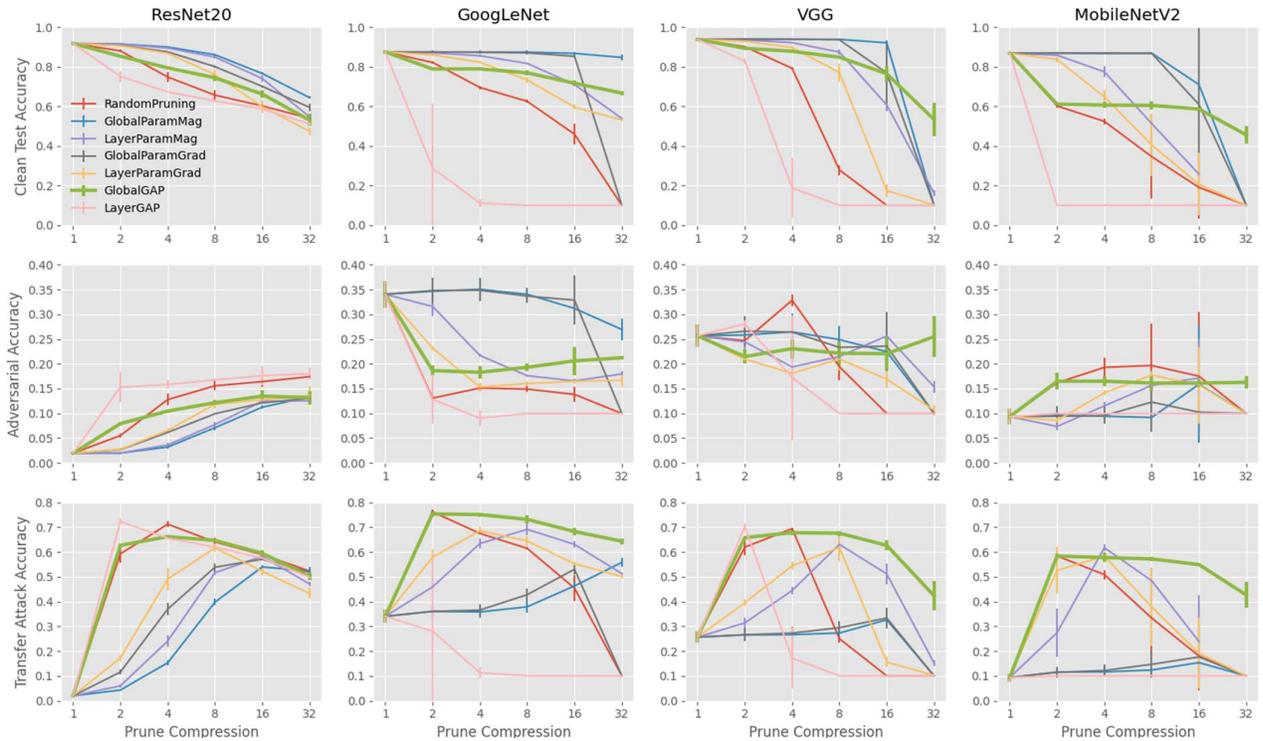

Fig. 2. Top1 test accuracy on clean inputs, Top1 adversarial accuracy, and Top1 transfer attack accuracy by model, pruning method, and compression ratio. Values are averages over three trials with error bars displaying one standard deviation.

In our experiments, we find that none of the pruning methods offer enough adversarial accuracy to be considered a defense. Further, there is not a pattern of how the compression ratio affects adversarial robustness, except for slightly more variability at higher compression. For ResNet, models become slightly more robust as compression increases, but for other architectures, robustness fluctuates. Pruning layerwise by gradient or parameter magnitude yields similar robustness, as does pruning globally by gradient or parameter magnitude. In the two instances that layerwise GAP learned, it achieve the highest adversarial robustness. MobileNetV2 and ResNet exhibit a trade-off between accuracy and robustness, which is supported by the findings in [9], but VGG and GoogLeNet show no such relationship. Finally, as compared to the transfer attack accuracy, models are most vulnerable against adversarial inputs generated for themselves, which is observed by [7].

The transfer attack accuracy shows the most interesting results. In almost all cases, the accuracy on adversarial inputs from the uncompressed model increases with pruning compression up to a certain point and then decreases. Global GAP is robust to transferred adversarial inputs for all compression ratios and for all models. Similar to its test accuracy, the GAP transfer attack accuracy decreases slower than other pruning methods as compression increases. Additionally, the GAP transfer attack accuracy is never far below its raw test accuracy. Random Pruning also achieves high transfer attack accuracy for low compression ratios, but loses both test and transferable accuracy quickly with higher compression. When layerwise GAP learned, it is usually the most robust to a transfer attack. Layer pruning in general has higher transfer accuracy than global pruning. For all conventional pruning methods, adversarial inputs transferred reasonably well from uncompressed to pruned models at low compression ratios, confirming the findings in [12].

While the models produced from GAP are still vulnerable to adversarial inputs generated on themselves, their robustness to adversarial inputs generated on their uncompressed counterparts suggests that GAP changes, but does not eliminate, the space of adversarial examples [4]. Other work, which found that stochastic pruning at inference time could improve accuracy on adversarial inputs [14], reinforces the idea that certain pruning methods are capable of changing the space of adversarial inputs. The high transfer attack accuracy of models produced from GAP shows that the decision boundaries around the adversarial inputs generated by the uncompressed model have shifted. These inputs are classified correctly far more often in the GAP model than in the unpruned model. In general, the results demonstrate the feasibility of increasing accuracy on a known set of adversarial inputs by removing a set of parameters which contribute their misclassification while mostly retaining accuracy on clean, unperturbed inputs.

An example of the distribution of weight values remaining after GAP and global gradient pruning is shown in Figure 1 for ResNet20 with pruning compression of 2 and the more extreme case of 32. The figure illustrates several findings. First, the values of the parameters which have the highest adversarial gradients (the difference between the uncompressed distribution and the GAP distribution) have a higher variance than parameters which have low clean gradients (those removed in global gradient pruning). Therefore, we observe that parameters with high adversarial gradients tend to be more evenly distributed across all possible parameter values than the distribution of parameters with high gradients on clean inputs.

Second, many more parameters with low magnitudes are kept, and many more with high magnitudes are removed, in GAP as compared to global gradient pruning. Also, while there is significant overlap in the parameters kept by both of these methods, each method also keeps many parameters that are pruned by the other method, causing disjointedness between the resulting models. The fact that both models can still learn the classification task might be explained by the hypothesis that overparameterized networks learn redundant features [24], or that some learned features in unpruned models help classify clean and adversarial inputs, while others only aid clean input classification [6], but further experimentation is necessary to solidify these hypotheses. Furthermore, the results suggest that there are partially disjoint subnetworks, such as the models generated from GAP and global gradient pruning, that can solve a classification task.

The effect of quantizing a model, pruned or unpruned, was negligible. Quantized models usually had both a small increase in adversarial robustness and nearly no change in clean or transfer accuracy over all compression ratios as also evidenced by [16]. These results may be explained by the error amplification effect [8], in which small perturbations may be muted, while large ones are amplified. The attack strength parameter $\epsilon = 8/255$ most likely falls into the former category.

## IV. CONCLUSION

In this work, we analyze the adversarial robustness of several pruning methods, as well as introduce and investigate a greedy adversarial pruning (GAP) algorithm that removes DNN parameters with the highest gradient on adversarial inputs. We find that while models generated from GAP do not achieve high enough adversarial accuracy to constitute a defense, they achieve high robustness to transfer attacks from their uncompressed counterparts, indicating that GAP has removed the model's tendency to misclassify these inputs by shifting classification boundaries.

## ACKNOWLEDGMENT

This work has been supported in part by a grant from the National Science Foundation.